\documentclass[letterpaper, 10 pt, journal, twoside]{ieeetran}
\usepackage{amsmath,amsfonts}
\usepackage{algorithmic}
\usepackage{algorithm}
\usepackage{array}
\usepackage[caption=false,font=normalsize,labelfont=sf,textfont=sf]{subfig}
\usepackage{textcomp}
\usepackage{stfloats}
\usepackage{url}
\usepackage{verbatim}
\usepackage{graphicx}
\usepackage{xcolor}
\usepackage{cite}
\usepackage{hyperref}
\usepackage{multirow}
\usepackage{bbding}
\usepackage{booktabs}
\usepackage{adjustbox}
\usepackage{threeparttable}
\usepackage{algorithm, algorithmic}
\usepackage{makecell}
\title{RadarGaussianDet3D: Gaussian Representation-based Real-time\\ 3D Object Detection with 4D Automotive Radars}

\author{Weiyi Xiong, Bing Zhu,~\IEEEmembership{Senior Member,~IEEE}, 
and Zewei Zheng,~\IEEEmembership{Member,~IEEE}%
\thanks{Corresponding Author: Zewei Zheng.}%
\thanks{Weiyi Xiong, Bing Zhu and Zewei Zheng are with the School of Automation Science and Electrical Engineering, Beihang University, Beijing, P.R.~China
(e-mail: weiyixiong@buaa.edu.cn; zhubing@buaa.edu.cn; zeweizheng@buaa.edu.cn).}%
\thanks{This paper has been accepted by IEEE Robotics and Automation Letters. Digital Object Identifier 	
10.1109/LRA.2026.3673988}
}

\begin{document}

\maketitle

\begin{abstract}
4D automotive radars have gained increasing attention for autonomous driving due to their low cost, robustness, and inherent velocity measurement capability.
However, existing 4D radar-based 3D detectors rely heavily on pillar encoders for BEV feature extraction, where each point contributes to only a single BEV grid, resulting in sparse feature maps and degraded representation quality.
In addition, they also optimize bounding box attributes independently, leading to sub-optimal detection accuracy. 
Moreover, their inference speed, while sufficient for high-end GPUs, may fail to meet the real-time requirement on vehicle-mounted embedded devices.
To overcome these limitations, an efficient and effective Gaussian-based 3D detector, namely RadarGaussianDet3D is introduced, leveraging Gaussian primitives and distributions as intermediate representations for radar points and bounding boxes.
In RadarGaussianDet3D, a novel Point Gaussian Encoder (PGE) is designed to transform each point into a Gaussian primitive after feature aggregation and employs the 3D Gaussian Splatting (3DGS) technique for BEV rasterization, yielding denser feature maps.
PGE exhibits exceptionally low latency, owing to the optimized algorithm for point feature aggregation and fast rendering of 3DGS.
In addition, a new Box Gaussian Loss (BGL) is proposed, which converts bounding boxes into 3D Gaussian distributions and measures their distance to enable more comprehensive and consistent optimization.
Extensive experiments on TJ4DRadSet and View-of-Delft demonstrate that RadarGaussianDet3D achieves high detection accuracy while delivering substantially faster inference, highlighting its potential for real-time deployment in autonomous driving. 
Source code is available at \href{https://github.com/XiongWeiyi/RadarGaussianDet3D}{https://github.com/XiongWeiyi/RadarGaussianDet3D}.
\end{abstract}

\vspace{-3mm}\section{Introduction}\label{intro}

\IEEEPARstart{A}{ccurate} and fast perception is essential for safety-critical autonomous driving. 
Since autonomous vehicles may operate under diverse conditions such as heavy rain or darkness, sensors capable of functioning reliably in all environments are necessary.
Unlike cameras and LiDARs, which rely on visible or invisible light, automotive radars employ millimeter waves that are robust against poor lighting and extreme weather, making them indispensable.
Furthermore, radars capture velocity information, which is crucial for understanding dynamic scenes.
Although conventional radars lack elevation measurement, preventing 3D environmental perception, this limitation has been overcome with the development of 4D radars.
In addition to enabling height sensing, 4D radars provide higher-resolution data, which enhances object identification and localization \cite{4d_radar_survey}.

Despite these advantages, 4D radars also suffer from clear drawbacks, namely data sparsity and noise. Compared with LiDARs, the resolution of 4D radars remains significantly lower \cite{4d_radar_survey}, resulting in sparse radar point clouds. Noise is also easily introduced due to factors such as the multi-path effect and receiver saturation \cite{RadarSplat}, making precise perception difficult.

\begin{figure}
    \centering
    \includegraphics[scale=0.43]{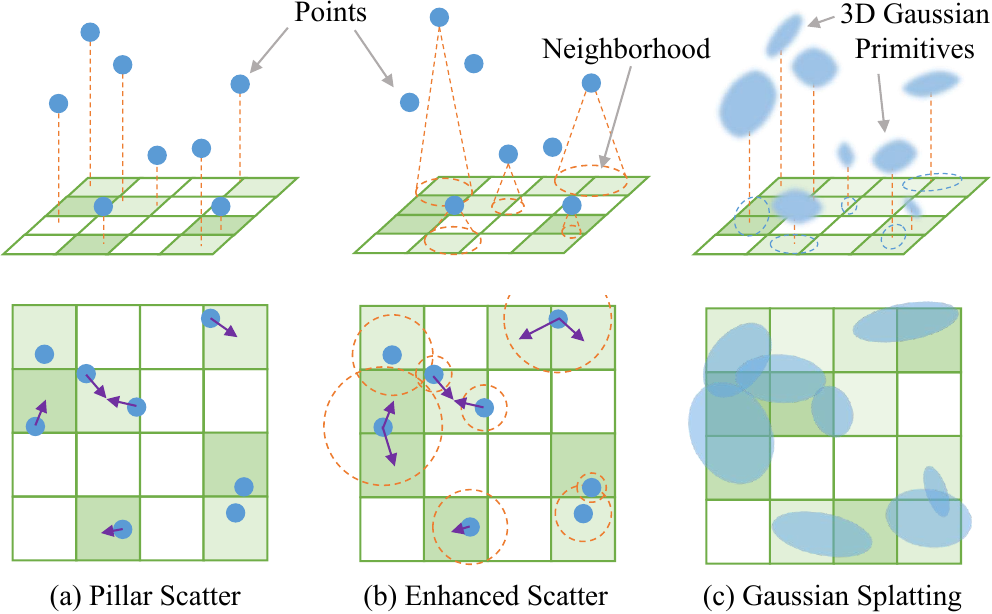}\vspace{-3mm}
    \caption{Illustration of different point scattering methods. Purple arrows from a point indicate the BEV grid cells influenced by that point. 
    (a) Pillar scatter in PointPillars \cite{PointPillars} maps each point to a single grid cell based on coordinates. 
    (b) Enhanced scatter methods, such as the RCS-aware scatter in RCBEVDet \cite{RCBEVDet}, define a neighborhood for each point and assign it to all grid cells whose centers lie within the neighborhood. 
    (c) Gaussian splatting in the proposed RadarGaussianDet3D rasterizes point-converted Gaussian primitives onto the BEV plane, allowing each point to contribute to all overlapping grid cells.}\label{fig:scatter}\vspace{-6mm}
\end{figure}

Within the perception framework, accurate and robust 3D object detection serves as a cornerstone for the subsequent trajectory prediction and motion planning in autonomous driving.
This paper addresses three major challenges encountered by existing 4D radar-based 3D detectors.
First, most detectors adopt the pillar encoder in PointPillars \cite{PointPillars} to construct radar bird’s-eye-view (BEV) feature maps.
As noted by Lin \textit{et al.} \cite{RCBEVDet} and shown in Fig.~\ref{fig:scatter}(a), the scatter operation in the pillar encoder maps sparse radar points to individual BEV grids based on their coordinates, producing sparse feature maps.
Although stacking additional BEV encoder layers can mitigate this problem, small object features may blend into background regions.
RCBEVDet \cite{RCBEVDet} proposes an enhanced scatter operation that assigns each point feature to all neighboring grids (Fig.~\ref{fig:scatter}(b)), with neighborhood size determined by the radar cross section (RCS), which roughly reflects object size.
However, this hand-crafted design is sub-optimal, since RCS is influenced not only by object size but also by factors such as material properties.
In addition, if a grid lies within the neighborhoods of multiple points, only the maximum feature value is retained, leading to information loss.
Therefore, a more suitable feature scattering strategy must be explored.

Second, existing bounding box regression losses (e.g., $L_1$ loss) typically measure the difference of each box attribute separately, neglecting correlations among them.
For instance, box localization error is usually computed without considering size or orientation, despite the fact that the same position error may have different effects depending on these factors.
Thus, a new loss function that jointly optimizes all bounding box attributes is required.

Third, autonomous vehicles must perceive and interpret their surroundings as quickly as possible to respond to potential hazards.
Although most 4D radar-based detectors achieve real-time performance on high-end GPUs, they may not meet real-time requirements on vehicle-mounted embedded devices with limited computational resources \cite{RadarPillars}, hindering deployment.
Hence, developing effective detectors with improved runtime performance remains imperative.

To address these issues, RadarGaussianDet3D is proposed as an efficient and effective 4D radar-based 3D object detector that employs Gaussian primitives and distributions as intermediate representations of radar points and bounding boxes.
For the first problem, the Point Gaussian Encoder (PGE) is designed to generate BEV feature maps using the 3D Gaussian Splatting (3DGS) technique \cite{3DGS}.
3DGS represents a scene with 3D Gaussian primitives, where a view is synthesized by projecting them onto a plane followed by differentiable rasterization.
Since a 3D Gaussian primitive can be regarded as a point with additional attributes such as size and rotation, it is natural to predict these attributes from radar points. 
Note that, unlike the original 3DGS framework that requires per-scene optimization, this process is feed-forward, single-shot, and generalizable.
The BEV feature map is then obtained through Gaussian splatting, which acts as a feature scattering strategy (Fig.~\ref{fig:scatter}(c)).
In this process, each point contributes to all grids overlapped by the projected 2D Gaussian primitive, and alpha-blending is used to integrate features from multiple points.
For bounding box regression, a Box Gaussian Loss (BGL) is proposed, which transforms predicted and ground-truth boxes into 3D Gaussian distributions and computes their distance, thereby comprehensively optimizing all box attributes.
Finally, since 3DGS \cite{3DGS} provides ultra-fast rendering and BGL is used only during training, the proposed model achieves high inference efficiency. 

The contributions of this work are summarized as follows:
\begin{itemize}
\item Radar points are modeled as 3D Gaussian primitives, and a Point Gaussian Encoder (PGE) is designed to efficiently splat these primitives into BEV feature maps, effectively alleviating the inherent sparsity of radar point clouds.
\item 3D bounding boxes are formulated as Gaussian distributions, and a Box Gaussian Loss (BGL) is introduced to measure box discrepancies via distributional distance, enabling joint optimization of all bounding box attributes.
\item By integrating PGE and BGL, a 4D radar-based 3D object detection framework, termed RadarGaussianDet3D, is developed. Experiments on the TJ4DRadSet\cite{TJ4DRadSet} and View-of-Delft\cite{VoD} datasets demonstrate a favorable accuracy–efficiency trade-off, achieving substantially lower latency than state-of-the-art methods while maintaining high detection accuracy.
\end{itemize}

This paper is organized as follows.
Section \ref{sec:related work} reviews related work, including 4D radar-based 3D object detection methods and applications of 3D Gaussian splatting in autonomous driving.
Section \ref{sec:method} presents the proposed RadarGaussianDet3D model in detail, and Section \ref{sec:experiments} reports and analyzes the experimental results.
Finally, Section \ref{sec:conclusion} concludes the paper.

\vspace{-3mm}\section{Related Work}\label{sec:related work}
\subsection{3D Object Detection with 4D Radar Point Clouds}
Represented as 3D point clouds, 4D radar data share certain characteristics with LiDAR data, and thus LiDAR-based detectors such as PointPillars \cite{PointPillars} can be directly applied \cite{VoD}.
However, this direct adaptation is sub-optimal because of the modality differences, requiring tailored modifications.

Several approaches focus on enriching radar features \cite{RPFA-Net,RCFusion,RadarPillars,SMURF}.
For instance, RCFusion \cite{RCFusion} processes spatial, velocity, and intensity values separately to avoid feature confusion.
RadarPillars \cite{RadarPillars} introduces self-attention among non-empty pillars to aggregate features belonging to the same object.
SMURF \cite{SMURF} predicts point density distributions using kernel density estimation as an additional feature, improving sparsity-awareness in detection.

Another line of work addresses the inherent deficiencies of 4D radar.
RCBEVDet \cite{RCBEVDet} defines a neighborhood for each point and scatters features across all BEV grids in the area, thereby densifying BEV feature maps and mitigating point cloud sparsity.
MAFF-Net \cite{MAFF-Net} proposes a cylindrical denoising assist module to identify keypoints around objects, reducing the effect of noise.

As certain limitations of 4D radar cannot be fully resolved algorithmically, some methods use auxiliary modalities during training.
E.g., SCKD \cite{SCKD} employs cross-modality distillation to transfer knowledge from LiDAR to 4D radar, enhancing performance without compromising efficiency.
Other approaches explore multi-modal fusion, integrating geometric information from LiDAR point clouds \cite{rl-fusion,L4DR,InterFusion} or semantic cues from RGB images \cite{LXL,LXLv2,SGDet3D,Doracamom} to complement 4D radar.

In contrast to the aforementioned methods, this work addresses the sparsity of 4D radar point clouds at the representation level. Instead of relying on the commonly used pillar-based representation, radar points are modeled as 3D Gaussian primitives, and these primitives are rasterized using 3DGS to generate denser BEV feature maps.

\begin{figure*}
    \centering
    \includegraphics[scale=0.4]{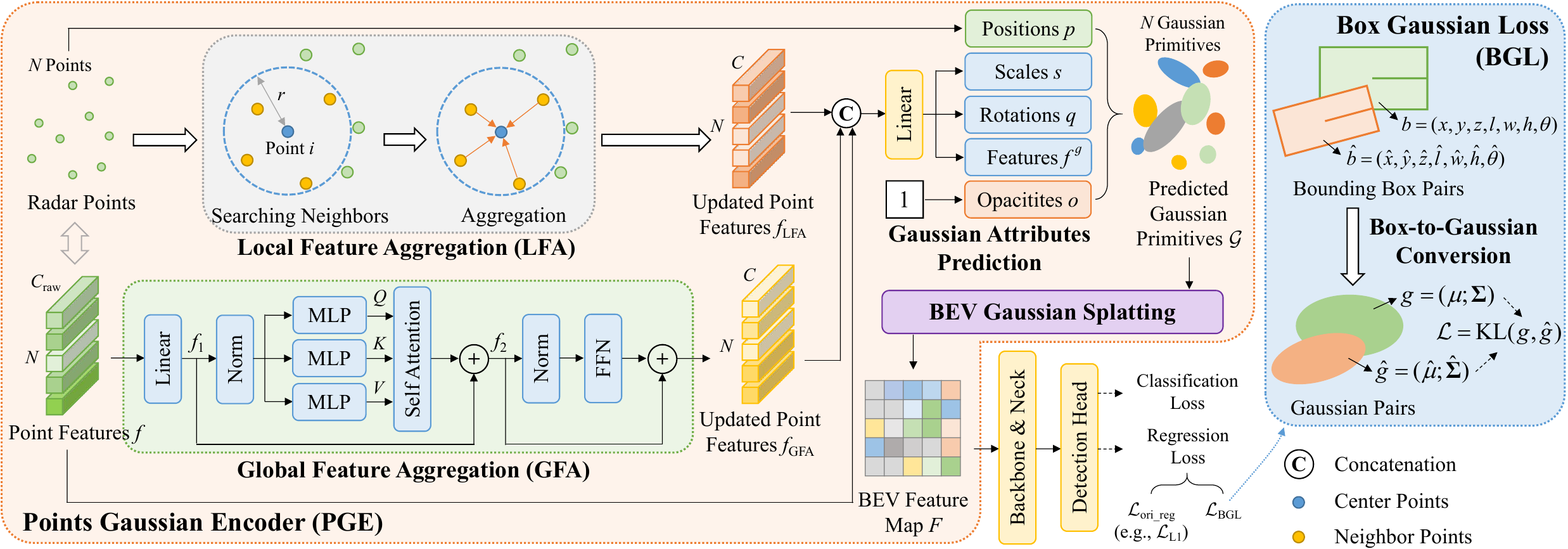}\vspace{-3mm}
    \caption{Overview of RadarGaussianDet3D.}
    \label{fig:overview}\vspace{-6mm}
\end{figure*}

\vspace{-3mm}\subsection{3D Gaussian Splatting in Autonomous Driving}

3D Gaussian Splatting (3DGS) \cite{3DGS} introduced a new paradigm in 3D reconstruction.
Although initially proposed for camera-based novel view synthesis (NVS) in static scenes, researchers extended it to other domains, including autonomous driving.
For instance, SplatAD \cite{SplatAD} and RadarSplat \cite{RadarSplat} extend 3DGS to NVS with LiDAR and radar sensors, respectively, while DrivingGaussian \cite{DrivingGaussian} adapts it to dynamic scenes.

3DGS has also been applied to various perception tasks, including BEV segmentation \cite{GaussianBeV,GaussianLSS}, 3D object detection \cite{RaGS}, and occupancy prediction \cite{GaussianFormer,GaussianFormer-2,GaussianWorld}.
However, perception tasks typically require strong generalization capability and real-time inference, making per-scene optimization of Gaussian attributes in the original 3DGS formulation impractical.
As a result, existing perception-oriented methods predict Gaussian attributes in a feed-forward manner, either directly from feature maps \cite{GaussianBeV,GaussianLSS,GaussianPretrain}  or via learnable queries \cite{GaussianFormer,GaussianFormer-2}.
In the feature-based paradigm, each Gaussian primitive is predicted directly from a feature map element (e.g., an image pixel), resulting in a simple and efficient design that is particularly suitable for single-modality inputs.
Moreover, this one-shot prediction incurs minimal computational overhead, making it highly favorable for real-time systems.
In contrast, query-based methods rely on iterative refinement of Gaussian primitives through repeated query-data interactions. 
While this design allows a variable number of Gaussian primitives and naturally supports multi-modal inputs, it introduces additional computational cost.
Beyond perception subtasks, query-based Gaussian representations have also been extended to end-to-end autonomous driving \cite{GaussianFusion,GaussianAD} and world models \cite{GaussianWorld}.

Once Gaussian primitives are predicted, they are splatted onto the BEV plane or rasterized in the 3D space, producing BEV pseudo-images or voxel grids that are subsequently processed by task-specific heads.

While existing approaches have achieved promising results, most focus on image-view transformation or multi-modal feature fusion. In contrast, this work pioneers the application of 3DGS to 4D radar point clouds for BEV feature extraction, where Gaussian attributes are directly predicted from sparse radar measurements. This design establishes a new paradigm for leveraging 3DGS in radar-based perception tasks.

\vspace{-3mm}\section{Proposed Method}\label{sec:method}\vspace{-1mm}
\subsection{Overview}
Fig.~\ref{fig:overview} presents the overall framework of the proposed RadarGaussianDet3D model.
Unlike most existing methods that rely on pillarization, RadarGaussianDet3D employs the concept of 3DGS to construct radar BEV feature maps through a dedicated Point Gaussian Encoder (PGE).
After aggregating both local and global features, the PGE predicts Gaussian attributes for each radar point, which are subsequently splatted onto the BEV plane.
The resulting feature map is then processed by the backbone, neck, and detection head to produce 3D bounding box predictions.

In addition, an auxiliary loss termed Box Gaussian Loss (BGL) is introduced during training.
Ground-truth and predicted bounding boxes are transformed into 3D Gaussian distributions, and their distributional distance is computed.
This formulation guides the network to comprehensively optimize all bounding box attributes in a unified manner.

\vspace{-3mm}
\subsection{Point Gaussian Encoder (PGE)}
The PGE consists of four main steps: Local Feature Aggregation (LFA), Global Feature Aggregation (GFA), Gaussian attribute prediction, and BEV Gaussian splatting.

\textbf{Local Feature Aggregation (LFA).}
The most straightforward approach to transform a point into a Gaussian primitive is to directly predict Gaussian attributes.
However, raw radar points contain limited information and are insufficient to support reliable prediction.
Therefore, local features are aggregated to enrich the representation of each point.

\begin{figure*}
    \centering
    \includegraphics[scale=0.4]{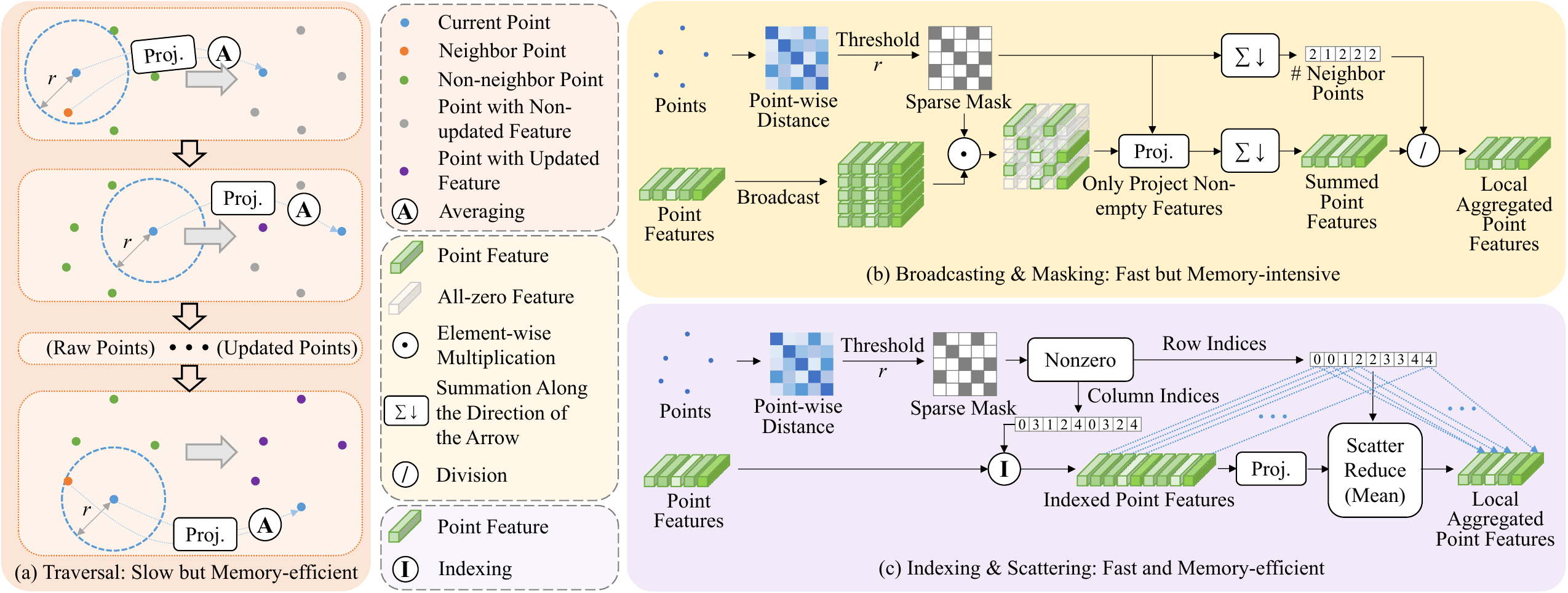}\vspace{-3mm}
    \caption{Visualization of different LFA implementations. 
    For simplicity, concatenation with position offsets is omitted.}
    \label{fig:lfa}\vspace{-6mm}
\end{figure*}

For the $i$-th point, a lightweight PointNet \cite{PointNet} is applied to all points in its spherical neighborhood $\mathcal N_i$:
\vspace{-1mm}\begin{equation}
    f^i_\textup{LFA}=\frac1{|\mathcal N_i|}\sum_{j\in\mathcal N_i}\mathtt{Linear}(\mathtt{Concat}([f_j,p_j-p_i])),
\end{equation}
where $p_i\in\mathbb R^3$ and $f_i\in\mathbb R^{C_\textup{raw}}$ denote the 3D coordinates and raw features of the $i$-th point $(i=1,\cdots,N)$, respectively.
Here, $f^i_\textup{LFA}\in\mathbb R^C$ represents the updated features after LFA.
$N$ denotes the total number of points, and $C$ refers to the channel dimension.
The $i$-th point is referred as the center point, while points in $\mathcal N_i$ are its neighbors.
In addition, the position offset from the center point to each neighbor is appended to features, following the pillar feature extraction in PointPillars \cite{PointPillars}.

To avoid traversing all points for averaging neighbor features and to reduce memory usage, an optimized algorithm is devised.
It relies on the PyTorch-supported operation $\mathtt{scatter\_reduce}$, as visualized in Fig.~\ref{fig:lfa}(c).
Experiments in Section~\ref{sec:ablation} confirm that the proposed Indexing \& Scattering algorithm achieves higher speed than the Traversal method in Fig.~\ref{fig:lfa}(a), while requiring considerably less memory compared with the Broadcasting \& Masking strategy in Fig.~\ref{fig:lfa}(b).

\textbf{Global Feature Aggregation (GFA).}
Previous studies \cite{MAFF-Net,RadarPillars} have shown that self-attention among non-empty pillars improves detection performance.
Inspired by this, self-attention is applied directly among points to capture global interactions in parallel with LFA.
Since radar points are sparse and most non-empty pillars contain only a single point, the computational complexity of point-based attention is comparable to that of pillar-based attention.

The GFA process is defined as Eq.~\eqref{eq:gfa}:
\begin{equation}\label{eq:gfa}
\begin{aligned}
    f_1&=\mathtt{Linear}(f),\\
    Q,K,V&=\mathtt{MLP}(\mathtt{LayerNorm}(f_1)),\\
    f_2&=\mathtt{SelfAttn}(Q,K,V)+f_1,\\
    f_\textup{GFA}&=\mathtt{FFN}(\mathtt{LayerNorm}(f_2))+f_2,
\end{aligned}
\end{equation}
where $f\in\mathbb R^{N\times {C_\textup{raw}}}$ denotes the raw point features, and $f_\textup{GFA}\in\mathbb R^{N\times C}$ represents the features after GFA.
Here, $\mathtt{SelfAttn}(Q,K,V)$ is the self-attention operation \cite{attention} with queries $Q$, keys $K$, and values $V$, while $\mathtt{FFN}(\cdot)$ denotes the feed-forward network.

\textbf{Gaussian Attribute Prediction.}
In the original 3DGS \cite{3DGS}, each Gaussian primitive is represented as $g=(\mu,s,q,o,sh)$, where $\mu\in\mathbb R^3$, $s\in\mathbb R^3$, $q\in\mathbb R^4$, $o\in\mathbb R$, and $sh\in\mathbb R^M$ denote the 3D position (mean), 3D scales, rotation quaternion, opacity, and spherical harmonic (SH) coefficients, respectively.
The SH coefficients are used to compute view-dependent colors $c$.
Since the objective in PGE is to obtain feature maps rather than images, the SH coefficients $sh$ are replaced with view-independent features $f^g\in\mathbb R^C$.
Moreover, the opacities and positions are not predicted.
Specifically, opacity $o$ is set to $1$, and the position offset is fixed to $0$ (i.e., $\mu=p$), as introducing excessive degrees of freedom for Gaussian primitives at an early stage may hinder learning.

Gaussian attributes for each radar point are obtained by concatenating the raw features with the enriched features produced by LFA and GFA, followed by a linear projection:
\begin{equation}
\begin{aligned}
    s',q,f^g&=\mathtt{Linear}(\mathtt{Concat}([f,f_\textup{LFA},f_\textup{GFA}])),\\
    s&=\mathtt{Sigmoid}(s') \cdot s_{\max},
\end{aligned}
\end{equation}
where $\mathtt{Sigmoid}$ ensures positive scale values while preventing excessive magnitude following\cite{TT-Occ}, and $s_{\max}=1\,\mathrm{m}$ is a predefined maximum scale. $q$ is normalized to satisfy the unit-norm constraint of quaternions.

Compared with original 3DGS paradigm, where Gaussian attributes are treated as learnable parameters optimized per scene, the proposed feed-forward prediction strategy ensures model generalizability across scenes. Moreover, the single-shot prediction is computationally more efficient than the iterative refinement in query-based Gaussian prediction methods.

\textbf{BEV Gaussian Splatting.}
In this step, the set of predicted Gaussian primitives $\mathcal G=\{g_i=(\mu_i,s_i,q_i,o_i,f^g_i)\}_{i=1}^N$ is splatted onto the BEV plane, with the implementation adapted from the CUDA code of 3DGS \cite{3DGS}.

Two main modifications are introduced in the 3D-to-2D projection and rendering.
First, unlike perspective projection onto the image plane, BEV projection is a parallel projection with scaling.
Given the target BEV resolution $(H,W)$ and the BEV range $(x,y)\in[x_{\min},x_{\max}]\times[y_{\min},y_{\max}]$, the 2D mean $\mu_\textup{2D}$ and 2D covariance matrix $\mathbf\Sigma_\textup{2D}$ are computed as
\begin{equation}
\begin{aligned}
    \mu_\textup{2D}&=\mathbf M\mu,\;\;\;\mathbf\Sigma_\textup{2D}=\mathbf M\mathbf\Sigma\mathbf M^\top,\\
    \mathbf M&=\begin{bmatrix}
        \frac W{x_{\max}-x_{\min}} & 0 & 0 \\
        0 & \frac H{y_{\max}-y_{\min}} & 0
    \end{bmatrix},
\end{aligned}
\end{equation}
where $\mathbf\Sigma=\mathbf R\mathbf S\mathbf S^\top\mathbf R^\top$, $\mathbf S=\mathtt{diag}(s)$, and $\mathbf R$ is the rotation matrix derived from quaternion $q$.

Second, the rendering equations of 3DGS \cite{3DGS} are modified by replacing the color $c$ with Gaussian features $f^g$, yielding the BEV feature map $F\in\mathbb R^{C\times H\times W}$:
\begin{equation}
\begin{aligned}
    F[\mathbf p]&=\sum_{i=1}^{N_\mathbf p}f^g_i\alpha_iT_i,\;\;\;T_i=\prod_{j=1}^{i-1}(1-\alpha_j),\\
    \alpha_i&=o_i\exp\left(-\frac12(\mathbf p-\mu_\textup{2D})^\top\mathbf\Sigma_\textup{2D}^{-1}(\mathbf p-\mu_\textup{2D})\right),
\end{aligned}
\end{equation}
where $F[\mathbf p]\in\mathbb R^C$ denotes the feature vector at pixel $\mathbf p$, and $N_{\mathbf p}$ is the number of Gaussian primitives overlapping with the tile containing $\mathbf p$.

Compared with the conventional pillar encoder that assigns each point to a single BEV grid, the proposed PGE allows each point to influence multiple neighboring grids adaptively, thereby generating denser BEV feature maps.

\vspace{-3mm}\subsection{Box Gaussian Loss (BGL)}
To measure differences between bounding boxes more comprehensively, we propose BGL, a loss function that represents 3D bounding boxes as 3D Gaussian distributions and computes the distance between them.

\textbf{Box-to-Gaussian Conversion.}
Since 3D bounding boxes and 3D Gaussian distributions share attributes ( e.g., position, scale, and orientation), they can be mutually converted.
Each ground-truth bounding box $b=[x,y,z,l,w,h,\theta]$ is transformed into 3D Gaussian distribution $g=\mathcal N(\mu;\mathbf\Sigma)$ as follows:
\begin{equation}
    \mu=[x,y,z],\;\;\; \mathbf\Sigma=\mathbf R\mathbf S\mathbf S^\top\mathbf R^\top,
\end{equation}
where
\begin{equation}
\begin{aligned}
    \mathbf S&=\mathtt{diag}([\frac l{2a},\frac w{2a},\frac h{2a}])=\frac1a\mathtt{diag}([\frac l2,\frac w2,\frac h2]),\\
    \mathbf R&=\begin{bmatrix}\cos\theta&-\sin\theta&0\\
    \sin\theta&\cos\theta&0\\
    0&0&1\end{bmatrix}
\end{aligned}
\end{equation}
with $\mathbf S$ and $\mathbf R$ denoting the scaling and rotation matrices, respectively, and $a>0$ as the scaling hyperparameter.

For a corresponding predicted bounding box $\hat b$, the same conversion yields a Gaussian distribution $\hat g=\mathcal N(\hat\mu;\mathbf{\hat\Sigma})$.

\begin{table*}[]
\centering
\caption{Results on TJ4DRadSet \texttt{test} set}\label{tab:tj}\vspace{-3mm}
\begin{threeparttable}[b]
\begin{tabular}{c|cccc|c|cccc|c|c}
\toprule
\multirow{2.5}{*}{Models} & \multicolumn{5}{c|}{3D AP (\%)} & \multicolumn{5}{c|}{BEV AP (\%)} & \multirow{2.5}{*}{\makecell[c]{FPS\\(Hz)}} \\ \cmidrule{2-11} 
 & Car & Ped. & Cyc. & Tru. & mAP & Car & Ped. & Cyc. & Tru. & mAP & \\ \midrule
PointPillars$^\ast$ (CVPR'19) \cite{PointPillars} & 21.26 & \underline{28.33} & 52.47 & 11.18 & 28.31 & 38.34 & \underline{32.26} & 56.11 & 18.19 & 36.23 & 42.9 \\
CenterPoint$^\ast$ (CVPR'21) \cite{CenterPoint} & 22.03 & 25.02 & 53.32 & 15.92 & 29.07 & 33.03 & 27.87 & \textit{58.74} & 26.09 & 36.18 & 34.5 \\
RPFA-Net$^\ast$ (ITSC'21) \cite{RPFA-Net} & \textit{26.89} & 27.36 & 50.95 & 14.46 & 29.91 & \underline{42.89} & 29.81 & 57.09 & 25.98 & 38.94 & - \\
PillarNeXt$^\ast$ (CVPR'23) \cite{PillarNeXt} & 22.33 & 23.48 & 53.01 & 17.99 & 29.20 & 36.84 & 25.17 & 57.07 & 23.76 & 35.71 & 28.0 \\
SMURF$^\ast$ (T-IV'23) \cite{SMURF} & \textbf{28.47} & 26.22 & \underline{54.61} & \textit{22.64} & \textit{32.99} & \textbf{43.13} & 29.19 & \underline{58.81} & \textit{32.80} & \textit{40.98} & 23.1\\
MUFASA (ICANN'24) \cite{MUFASA} & 23.56 & 23.70 & 48.39 & \underline{25.25} & 30.23 & 41.25 & 24.54 & 53.64 & \textbf{36.97} & 39.10 & - \\
MAFF-Net (RAL'25) \cite{MAFF-Net} & \underline{27.31} & \textbf{33.13} & \textit{54.35} & \textbf{26.71} & \textbf{35.38} & 39.05 & \textbf{35.25} & 56.35 & \underline{35.73} & \underline{41.59} & 17.9$^\dag$ \\ \midrule
RadarGaussianDet3D \textbf{(Ours)} & 26.69 & \textit{28.18} & \textbf{65.84} & 19.63 & \underline{35.08} & \textit{41.66} & \textit{30.28} & \textbf{69.62} & 26.35 & \textbf{41.98} & \textbf{43.5} \\ \bottomrule
\end{tabular}
\begin{tablenotes}
    \item[1] In each column, the highest value is in \textbf{bold}, the 2nd highest \underline{underlined}, and the 3rd highest in \textit{italic}.
    \item[2] Inference speeds marked with $\dag$ are measured on RTX 4090, while others are on Tesla V100.
    \item[3] The results of models marked with $\ast$ are from \cite{SMURF}, while others from corresponding citations.
\end{tablenotes}
\end{threeparttable}\vspace{-7mm}
\end{table*}

\textbf{Box Gaussian Loss.}
Given two Gaussian distributions, their distance is measured as BGL.
In this work, the KL divergence is adopted as the distance metric:
\vspace{-2mm}\begin{equation}
    \mathcal L_\text{BGL}=\frac1{N_b}\sum_{i=1}^{N_b} \textup{KL}(\hat g_i, g_i),
\end{equation}
where $N_b$ denotes the number of ground-truth bounding boxes; %and
\vspace{-2mm}\begin{equation}\label{klloss}
    \textup{KL}(\hat g,g)=\frac12[(\hat\mu-\mu)^\top{\mathbf\Sigma}^{-1}(\hat\mu-\mu)+\textup{Tr}(\mathbf\Sigma^{-1}\hat{\mathbf\Sigma})+\log\frac{|\mathbf\Sigma|}{|\hat{\mathbf\Sigma}|}-3]
\end{equation}
represents the KL divergence between the predicted Gaussian distribution $\hat g$ and the ground-truth one $g$. Unlike $\mathrm{KL}(g,\hat g)$, the $\mathbf{\Sigma}^{-1}$ term in the first term of $\mathrm{KL}(\hat g, g)$ remains constant during training, leading to more stable optimization of the Gaussian mean. In addition, to ensure numerical stability during covariance inversion, the predicted bounding box scales are clamped to a small positive lower bound.

The first term $(\hat\mu-\mu)^\top\mathbf\Sigma^{-1}(\hat\mu-\mu)$ is the Mahalanobis distance between $\hat\mu$ and $\mathcal N(\mu;\mathbf\Sigma)$.
Unlike $L_1$ distance, it accounts for the shape and orientation of the ground-truth object, thereby providing more informative localization error.
The remaining terms capture differences in scales and orientations.

It should be noted that the hyperparameter $a$ influences only the first term, while its effects in the other terms are canceled through division, making the result independent of $a$.
For larger objects, $a$ should be set higher, since large Euclidean distances may otherwise yield small Mahalanobis distances.

The total regression loss $\mathcal L_\text{reg}$ is defined as
\vspace{-2mm}\begin{equation}
    \mathcal L_\text{reg}=\mathcal L_\text{ori\_reg}+\lambda\mathcal L_\text{BGL},
\end{equation}
where $\mathcal L_\text{ori\_reg}$ denotes the original regression loss used in the detection head (e.g., $L_1$ loss in CenterHead \cite{CenterPoint}), and $\lambda$ is the balancing weight.
The classification loss remains unchanged.

\vspace{-3mm}\section{Experiments}\label{sec:experiments}
\subsection{Implementation Details}
\textbf{Datasets and Evaluation Metrics.}
Two publicly available datasets are used to evaluate the proposed RadarGaussianDet3D: TJ4DRadSet \cite{TJ4DRadSet} and View-of-Delft (VoD) \cite{VoD}.
Both datasets provide synchronized data from a 4D radar, a camera, and a LiDAR, with tens of thousands of annotated 3D bounding boxes.
Compared with VoD, TJ4DRadSet is collected under more complex road and traffic conditions with greater lighting variability, making detection significantly more challenging. Experiments on both TJ4DRadSet and VoD are conducted using their official dataset partitioning, with models trained and evaluated on each dataset independently.

Following the official TJ4DRadSet protocol, both 3D and BEV average precisions (APs) are evaluated over the region
$D=[x_{\min},x_{\max}]\times[y_{\min},y_{\max}]\times[z_{\min},z_{\max}]=[0,69.12\mathrm m]\times[-39.68\mathrm m,39.68\mathrm m]\times[-4\mathrm m,2\mathrm m]$
for car, pedestrian, cyclist, and truck.

For the VoD dataset, the 3D AP is computed for the categories of car, pedestrian, and cyclist in both the entire annotated area ($D_\textup{EAA}$) and the region of interest ($D_\textup{ROI}$):
\vspace{-2mm}\begin{equation}
    \begin{aligned}
    D_\textup{EAA}=&[x_{\min},x_{\max}]\times[y_{\min},y_{\max}]\times[z_{\min},z_{\max}]\\=&[0,51.2\mathrm m]\times[-25.6\mathrm m,25.6\mathrm m]\times[-3\mathrm m,2\mathrm m],\\
    D_\textup{ROI}=&[x^c_{\min},x^c_{\max}]\times[z^c_{\min},z^c_{\max}]\\
    =&[-4\mathrm m,4\mathrm m]\times[0\mathrm m,25\mathrm m],
    \end{aligned}
\end{equation}
where ROI is defined in the camera coordinate system, and the coordinates $x,y,z$ carry the superscript $c$.

\begin{table*}[]
\centering
\caption{Results on VoD \texttt{val} set}\label{tab:vod}\vspace{-3mm}
\begin{threeparttable}[b]
\begin{tabular}{cc|ccc|c|ccc|c|cccc}
\toprule
\multirow{2.5}{*}{Models} & \multirow{2.5}{*}{Modality} & \multicolumn{4}{c|}{EAA AP (\%)} & \multicolumn{4}{c|}{ROI AP (\%)} & \multicolumn{4}{c}{FPS (Hz)} \\ \cmidrule{3-14} 
 &  & Car & Ped. & Cyc. & mAP & Car & Ped. & Cyc. & mAP & V100 & 3090 & 4090 & Xavier\\ \midrule
PointPillars$^\diamond$ (CVPR'19) \cite{PointPillars} & R & 38.8 & 34.4 & 66.9 & 46.7 & 71.9 & 45.1 & \textit{88.4} & 67.8 & 78.4 & 178.4 & 187.0$^\star$ & 20.6 \\
CenterPoint$^\star$ (CVPR'21) \cite{CenterPoint} & R & 32.7 & 38.0 & 65.5 & 45.4 & 62.0 & 48.2 & 85.0 & 65.1 & - & - & 72.2& - \\
PillarNeXt$^\star$ (CVPR'23) \cite{PillarNeXt} & R & 30.8 & 33.1 & 62.8 & 42.2 & 66.7 & 39.0 & 85.1 & 63.6 & - & - & 67.2 & - \\
SMURF (T-IV'23) \cite{SMURF} & R & \underline{42.3} & 39.1 & 71.5 & 51.0 & 71.7 & 50.5 & 86.9 & 69.7 & 30.0 & - & - & -\\
RadarPillars$^\diamond$ (ITSC'24) \cite{RadarPillars} & R & 41.1 & 38.6 & \textit{72.6} & 50.7 & 71.1 & \underline{52.3} & 87.9 & \textit{70.5} & 82.8 & 179.1 & - & 34.4\\
MUFASA (ICANN'24) \cite{MUFASA} & R & \textbf{43.1} & 39.0 & 68.7 & 50.2 & \textbf{72.5} & 50.3 & \underline{88.5} & 70.4 & - & - & - & - \\
4DRadDet (ICRA'25) \cite{4DRadDet} & R & 42.0 & \textit{40.7} & 71.6 & \textit{51.4} & \textit{72.1} & 51.2 & 88.0 & 70.4 & - & - & 35.2 & - \\
MAFF-Net$^\star$ (RAL'25) \cite{MAFF-Net} & R & \underline{42.3} & \textbf{46.8} & \textbf{74.7} & \textbf{54.6} & \underline{72.3} & \textbf{57.8} & 87.4 & \textbf{72.5} & - & - & 28.7 & - \\ \midrule
RadarGaussianDet3D \textbf{(Ours)} & R & 40.7 & \underline{42.4} & \underline{73.0} & \underline{52.0} & 71.2 & \textit{51.7} & \textbf{89.0} & \underline{70.6} & \textbf{83.2} & - & - & - \\ \midrule\midrule
SCKD (AAAI'25) \cite{SCKD} & R+(L) & 41.9 & 43.5 & 70.8 & 52.1 & 77.5 & 51.1 & 86.9 & 71.8 & - & - & 39.3 & -\\
\bottomrule
\end{tabular}
\begin{tablenotes}
    \item[1] In the column of Modality, R denotes radar, and (L) represents that LiDAR is incorporated during training.
    \item[2] In each column, the highest value is in \textbf{bold}, the 2nd highest \underline{underlined}, and the 3rd highest in \textit{italic}.
    \item[3] Inference speeds are measured on Tesla V100, RTX 3090, RTX 4090 %(mainstream GPUs) 
    or AGX Xavier (an embedded device designed for autonomous driving).
    \item[4] The results of models / inference speeds marked with $\star$/$\diamond$ are from \cite{MAFF-Net} / \cite{RadarPillars}, while others from corresponding citations.
\end{tablenotes}
\end{threeparttable}\vspace{-7mm}
\end{table*}

\textbf{Network and Hyper-parameters.}
The proposed model is developed based on CenterPoint-Pillar \cite{CenterPoint}, with the pillar encoder replaced by the proposed PGE and the BGL incorporated.
In BEV Gaussian splatting, the target feature map size $(H,W)$ is set to $(320,320)$ for VoD and $(432,496)$ for TJ4DRadSet, ensuring the same resolution as the output of the pillar encoder with the commonly adopted pillar size of $0.16 \mathrm m$ \cite{SMURF,RadarPillars,MAFF-Net}, for fair comparison.
The spherical neighborhood radius in LFA is set to $r=0.32\mathrm m$, where small deviations have only a marginal impact on detection accuracy, and the feature dimension in PGE is fixed to $C=64$.

During training, the scaling factor $a$ in BGL is set to $1$ for pedestrians and cyclists, and to $3$ for cars and trucks.
The loss weight for BGL is fixed to $1.0$.

\textbf{Training Details.}
The proposed model is implemented using the MMDetection3D framework \cite{mmdet3d} and trained on a single NVIDIA Tesla V100 GPU for 24 epochs with a batch size of 8.
AdamW is employed as the optimizer, with an initial learning rate of $2\times 10^{-4}$ scheduled by cosine annealing.

\vspace{-3mm}\subsection{Comparison with State-of-the-Arts}
\textbf{Results on TJ4DRadSet.}
Table~\ref{tab:tj} reports the detection accuracy and inference speed of different models on TJ4DRadSet \cite{TJ4DRadSet}, including the proposed RadarGaussianDet3D.
It achieves accuracy comparable to the state-of-the-art MAFF-Net \cite{MAFF-Net} ($-0.3\%~\text{mAP}_\text{3D}, +0.4\%~\text{mAP}_\text{BEV}$), while delivering substantially higher inference speed.
This efficiency is maintained even on a lower-performance GPU, leaving room for future integration of more computationally intensive enhancements.

\begin{figure}
    \centering
    \includegraphics[scale=0.88]{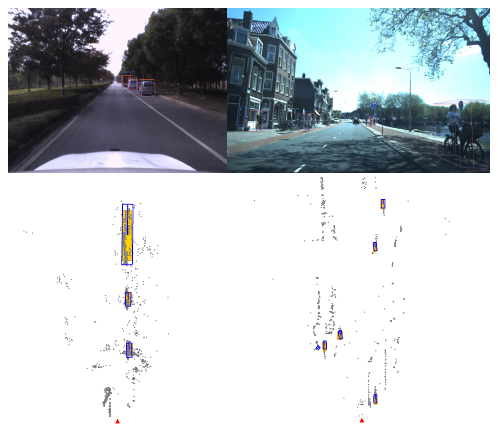}\vspace{-4mm}
    \caption{Visualization results of RadarGaussianDet3D on the TJ4DRadSet \cite{TJ4DRadSet} \texttt{test} set (left) and VoD \texttt{val} set (right).
    Gray points denote 4D radar points in BEV, orange and blue boxes indicate ground-truth and predicted bounding boxes, respectively, and the red triangle marks the ego-vehicle position.}
    \label{fig:vis}\vspace{-5mm}
\end{figure}

\textbf{Results on VoD.}
To further assess generalizability, experiments are conducted on the VoD \cite{VoD} dataset, with results summarized in Table~\ref{tab:vod}.
RadarGaussianDet3D ranks second among models with radar single-modality, outperforming most methods and only falling behind the two-stage detector MAFF-Net \cite{MAFF-Net}.
Notably, the performance gap between RadarGaussianDet3D and SCKD \cite{SCKD}, which leverages both LiDAR and radar during training, is merely $0.1\%~\text{mAP}_\text{EAA}$, highlighting the effectiveness of the proposed approach.

\textbf{Inference Speed Considerations.}
Musiat \textit{et al.} \cite{RadarPillars} have benchmarked several models (including PointPillars \cite{PointPillars}) across GPUs and embedded devices on the VoD \cite{VoD} dataset, which enables fairer comparison of inference speeds of different models.
From Table \ref{tab:vod}, it can be inferred that models with frame rates significantly below that of PointPillars \cite{PointPillars} are unlikely to achieve real-time performance (i.e., FPS $>10 \mathrm{Hz}$) on embedded platforms such as NVIDIA AGX Xavier.
Consequently, although MAFF-Net \cite{MAFF-Net} yields the highest accuracy, it is unlikely to meet real-time requirements in practical deployments.
In contrast, RadarGaussianDet3D operates even faster than PointPillars \cite{PointPillars}, suggesting strong potential for real-time deployment on embedded devices in autonomous driving systems. The speed gain of PGE primarily stems from avoiding point padding and redundant computation on empty points in pillar-based encoders, together with the efficiency of Gaussian splatting.

\vspace{-3mm}\subsection{Ablation Study}\label{sec:ablation}
This section presents ablation studies to evaluate the effectiveness of the proposed components. Results on TJ4DRadSet \cite{TJ4DRadSet} dataset are reported by default.

\textbf{Ablations on PGE and BGL.}
Table~\ref{tab:ablation1} reports the performance of models obtained by gradually adding modules from the CenterPoint-Pillar \cite{CenterPoint} baseline (a) to the complete RadarGaussianDet3D (g).
Replacing the pillar encoder with Gaussian splatting increases the 3D mAP by $0.1\%$ on TJ4DRadSet, since each point contributes to multiple BEV grids, yielding denser feature maps.
Adding LFA/GFA brings additional gains of $1.3\%$ and $1.5\%~\text{mAP}\text{3D}$, respectively, demonstrating that point-level feature interaction benefits Gaussian attribute prediction.
When combined to form the proposed PGE, the performance reaches $33.50\%~\text{mAP}_\text{3D}$, indicating that LFA and GFA capture complementary information.
Finally, the inclusion of BGL further improves accuracy by $1.6\%$, attributable to its comprehensive measurement of bounding box differences.

On VoD \cite{VoD}, a similar overall performance trend is observed, except that detection accuracy decreases from (a) to (b) due to the sparser and more irregular radar point distributions. This observation emphasizes the necessity of the proposed LFA and GFA modules, which effectively recover and further improve performance by aggregating local and global point features.

\begin{table*}[]
\centering
\caption{Ablation Studies of Components on TJ4DRadSet \texttt{test} set and VoD \cite{VoD} \texttt{val} set}\label{tab:ablation1}\vspace{-3mm}
\begin{tabular}{c|cccc||cccc|c|cccc|c||cc}
\toprule
\multirow{2.5}{*}{} & \multicolumn{4}{c||}{Enhancements} & \multicolumn{5}{c|}{TJ4DRadSet 3D AP (\%)} & \multicolumn{5}{c||}{TJ4DRadSet BEV AP (\%)} &  \multicolumn{2}{c}{VoD mAP (\%)} \\ \cmidrule{2-17} 
 & GS & LFA & GFA & BGL & Car & Ped. & Cyc. & Tru. & mAP & Car & Ped. & Cyc. & Tru. & mAP & EAA & ROI \\ \midrule
(a) &  &  &  &  & 23.01 & 23.32 & 57.58 & 19.25 & 30.79 & 35.82 & 26.25 & 62.65 & 27.95 & 38.42 & 46.84  & 68.51\\
(b) & \Checkmark &  &  &  & 23.05 & 23.76 & 57.45 & 19.30 & 30.89 & 36.11 & 27.63 & 61.90 & 28.71 & 38.59 & 46.06  & 67.00\\
(c) & \Checkmark & \Checkmark &  &  & 22.99 & 27.94 & 61.50 & 16.39 & 32.20 & 39.53 & 30.20 & 64.84 & \textbf{29.45} & 41.00 & 48.70  & 69.12\\
(d) & \Checkmark &  & \Checkmark &  & 22.80 & 27.04 & 61.83 & 17.63 & 32.33 & 34.89 & 28.96 & 67.22 & 26.81 & 39.47 & 48.94 & 69.48\\
(e) & \Checkmark & \Checkmark & \Checkmark &  & 25.77 & 28.08 & 64.55 & 15.61 & 33.50 & 39.34 & \textbf{30.57} & 68.30 & 26.41 & 41.15 & 50.18  & 69.52\\
(f) &  &  &  & \Checkmark & 25.77 & 23.47 & 58.42 & 18.94 & 31.65 & \textbf{42.44} & 26.60 & 63.04 & 28.20 & 40.07 & 48.30  & 69.52\\
(g) & \Checkmark & \Checkmark & \Checkmark & \Checkmark & \textbf{26.69} & \textbf{28.18} & \textbf{65.84} & \textbf{19.63} & \textbf{35.08} & 41.66 & 30.28 & \textbf{69.62} & 26.35 & \textbf{41.98} & \textbf{52.04}  & \textbf{70.61}\\ \bottomrule
\end{tabular}\vspace{-6mm}
\end{table*}

\begin{figure}
    \centering
    \includegraphics[scale=0.88]{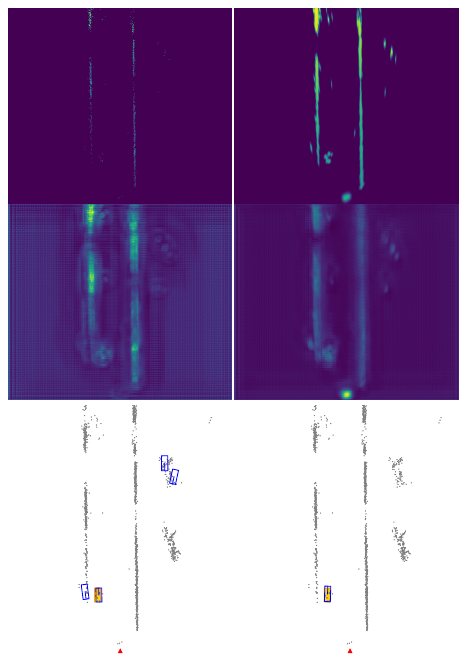}\vspace{-2mm}
    \caption{Visualization of BEV feature maps from CenterPoint-Pillar \cite{CenterPoint} (left) and RadarGaussianDet3D without BGL (right) on the TJ4DRadSet \texttt{test} set. 
    The first row presents initial BEV feature map after the pillar encoder or PGE, the second row displays the BEV feature map output by the backbone and neck, and the last row shows the radar point cloud and final detection results.} 
    \label{fig:bev_feat_vis}\vspace{-8mm}
\end{figure}

\textbf{Ablations on Predicted Gaussian Attributes in PGE.}
Table~\ref{tab:ablation2} presents results for different predicted Gaussian attributes, with all models trained without BGL.
Predicting position offsets and opacities leads to performance degradation, confirming that excessive degrees of freedom for Gaussian primitives from sparse radar points are detrimental, likely due to insufficient local context and measurement noise inherent to the radar modality. Optimal performance is obtained when the model focuses on estimating scale and rotation attributes.

\textbf{BEV Feature Map Visualization.}
To illustrate the advantages of PGE, BEV feature maps from CenterPoint-Pillar \cite{CenterPoint} and RadarGaussianDet3D (without BGL) are compared.
The only difference between these models lies in the module used to transform radar points into BEV feature maps (pillar encoder vs. PGE).
As shown in Fig.~\ref{fig:bev_feat_vis}, PGE produces feature maps with broader activation, capturing continuous road boundaries and clearer foreground objects.
After processed by the backbone and neck, the pillar-based feature map becomes noisy and over-activated, a consequence of hallucinations when completing sparse maps.
This effect is notably reduced in the PGE output, leading to more accurate detections.

\begin{table}[]
    \centering
    \caption{Ablation Studies of Predicted Gaussian Attributes}\label{tab:ablation2}\vspace{-3mm}
    \begin{tabular}{c|cc|cc}
    \toprule
    \multirow{2}{*}{} & \multicolumn{2}{c|}{Predicted Attributes} & \multirow{2.5}{*}{3D AP (\%)} & \multirow{2.5}{*}{BEV AP (\%)} \\ \cmidrule{2-3}
     & Position Offset & Opacity &  &  \\  \midrule
    (a) & \Checkmark & \Checkmark & 32.39 & 39.72 \\
    (b) &  & \Checkmark & 33.30 & 40.14 \\ 
    (c) &  &  & \textbf{33.50} & \textbf{41.15} \\ \bottomrule
    \end{tabular}\vspace{-3mm}
\end{table}

\begin{table}[]
    \centering
    \caption{The complexity of different LFA implementations}
    \label{tab:lfa_complexity}\vspace{-3mm}
    \begin{tabular}{c|rr}
    \toprule
    Algorithm & Avg. Runtime & Max. Memory \\ \midrule
    Traversal & 177.9ms & \textbf{202.6MB} \\
    Broadcasting \& Masking & 3.9ms & 3981.6MB \\
    Indexing \& Scattering & \textbf{0.5ms} & \textbf{202.6MB} \\ \bottomrule
    \end{tabular}\vspace{-6mm}
\end{table}

\textbf{Complexity of Different LFA Implementations.}
Table~\ref{tab:lfa_complexity} compares the runtime and peak memory consumption of the LFA implementations in Fig.~\ref{fig:lfa}, evaluated on the TJ4DRadSet \cite{TJ4DRadSet} \texttt{test} set using an NVIDIA Tesla V100 GPU.
As expected, the Traversal method is the slowest.
Broadcasting \& Masking benefits from PyTorch’s optimized matrix operations and achieves faster runtime, but at the cost of sharply increased memory usage due to repeated point features.
In contrast, the proposed Indexing \& Scattering algorithm leverages mask sparsity to avoid redundant computation and storage, maintaining low memory consumption while further improving speed.

\vspace{-3mm}\section{Conclusion}\label{sec:conclusion}
This paper presented RadarGaussianDet3D, a fast and accurate 4D radar-based 3D object detector.
To mitigate the inherent sparsity of radar point clouds, the proposed Point Gaussian Encoder (PGE) replaces the conventional pillar encoder by predicting Gaussian primitives from aggregated point features and applying 3D Gaussian Splatting to generate dense BEV feature maps.
In addition, the Box Gaussian Loss (BGL) comprehensively measures bounding box differences by accounting for correlations among attributes through box-to-Gaussian conversion.
Extensive experiments demonstrated that RadarGaussianDet3D achieves high accuracy while delivering substantially lower inference latency, highlighting its suitability for real-time autonomous driving applications.

By unifying radar points and 3D bounding boxes under Gaussian-based representations, this work contributes new perspectives to 4D radar-based 3D object detection.
Future research will extend this paradigm to multi-modal fusion and temporal modeling for perception in complex scenarios.

\bibliographystyle{IEEEtran}
\bibliography{reference}
\end{document}